# Predicting adverse outcomes following catheter ablation treatment for atrial fibrillation

**Juan C. Quiroz[1] PhD, David Brieger[2,3] PhD, Louisa Jorm[1] PhD, Raymond W Sy[2,3] PhD, Benjumin Hsu[1] PhD, Blanca Gallego[1] PhD**

[1]Centre for Big Data Research in Health, University of New South Wales, Sydney, Australia

[2]Department of Cardiology, Concord Repatriation General Hospital, Sydney, Australia

[3]Faculty of Medicine and Health, University of Sydney, Sydney, Australia

**\* Correspondence:**
Juan C. Quiroz
juan.quiroz@unsw.edu.au

## Abstract

**Objective:** To develop prognostic survival models for predicting adverse outcomes after catheter ablation treatment for non-valvular atrial fibrillation (AF).

**Methods:** We used a linked dataset including hospital administrative data, prescription medicine claims, emergency department presentations, and death registrations of patients in New South Wales, Australia. The cohort included patients who received catheter ablation for AF. Traditional and deep survival models were trained to predict major bleeding events and a composite of heart failure, stroke, cardiac arrest, and death.

**Results:** Out of a total of 3285 patients in the cohort, 177 (5.3%) experienced the composite outcome—heart failure, stroke, cardiac arrest, death—and 167 (5.1%) experienced major bleeding events after catheter ablation treatment. Models predicting the composite outcome had high risk discrimination accuracy, with the best model having a concordance index > 0.79 at the evaluated time horizons. Models for predicting major bleeding events had poor risk discrimination performance, with all models having a concordance index < 0.66. The most impactful features for the models predicting higher risk were comorbidities indicative of poor health, older age, and therapies commonly used in sicker patients to treat heart failure and AF.

**Discussion:** Diagnosis and medication history did not contain sufficient information for precise risk prediction of experiencing major bleeding events. Predicting the composite outcome yielded promising results, but future research is needed to validate the usefulness of these models in clinical practice.

**Conclusions:** Machine learning models for predicting the composite outcome has the potential to enable clinicians to identify and manage high-risk patients following catheter ablation proactively.

# 1 Introduction

Atrial fibrillation (AF) is the most common cardiac arrhythmia. AF is a significant driver of cardiovascular hospitalization, and it is associated with adverse outcomes including stroke, heart failure, and mortality (1,2). Studies have shown worldwide increasing trends in AF (3,4), which makes the treatment of AF a significant target for managing patients' cardiac health and reducing cardiac-related deaths. Amongst the treatment options recommended for managing AF (5,6), catheter ablation has been increasingly used and continues to be the focus of comparative effectiveness research. Recent clinical trials have demonstrated that compared to medical therapy, catheter ablation can lead to better outcomes in selected populations (7–9).

While machine learning (ML) has been increasingly applied in studies of diagnosis, risk prediction, and management of AF, only a few studies have focused on using ML for prognostic modeling of adverse outcomes following catheter ablation treatment (10,11). AF recurrence following ablation has been the adverse outcome most commonly analyzed in prior studies, with ML models trained with laboratory and clinical parameters (12) and magnetic resonance imaging scans (13,14). One study used ML to predict non-pulmonary vein trigger origins to reduce AF recurrence after ablation (15). Other studies used ML to predict 30-day and 90-day readmission following catheter ablation (16,17). One study predicted major cardiac outcomes (major bleeding, stroke/systemic embolism, and death) with ML, but the cohort was newly diagnosed AF patients treated with vitamin K antagonists and not catheter ablation (18). Beyond ML, statistical risk scores are the models most commonly used for predicting risk of adverse outcomes for AF patients (e.g. the $CHADS_2$ score for stroke prediction) and recurrence of AF after catheter ablation (19). The major limitation of these ML studies for predicting AF ablation outcomes is that the models predict binary outcomes (probability of experiencing the event), whereas a time-to-event modeling approach would enable the calculation of survival curves and risk estimates at various points of interest. A survival time-to-event approach also allows less biased modeling in the presence of loss to follow-up and competing events.

Given the 20-40% recurrence of AF in patients who undergo ablation and the association of AF with adverse outcomes (5,20), prognostic models could enable clinicians to develop treatment and management plans for patients, and act as a communication tool with patients who can be part of the decision-making process by knowing their risk profile. To date, there are no ML survival prognostic models for predicting the risk of experiencing major cardiac adverse outcomes after catheter ablation. This study aims to develop prognostic ML models that predict risk of experiencing adverse outcomes, including major bleeding, heart failure, stroke, cardiac arrest, and death following catheter ablation.

# 2 Materials and methods

## 2.1 Data sources

This study used linked administrative inpatient and emergency department data, pharmaceutical claims data, and mortality data for patients from New South Wales (NSW), Australia. Data on AF hospitalizations were extracted from the NSW Admitted Patient Data Collection (APDC), which includes records of all inpatient separations (discharges, transfers, and deaths) from public and private hospitals in NSW (see Appendix 1 for details). Data on emergency visits were extracted from the NSW Emergency Department Data Collection (EDDC), which includes records of presentations to public hospital emergency departments (Eds) in NSW. Pharmaceutical dispensing data were obtained from Pharmaceutical Benefits Scheme (PBS) records, which contain claims for subsidized prescription medicines in Australia dispensed in community pharmacies and private hospitals. Prescription medicines dispensed in NSW public hospitals are not included in PBS data (21). Death records were extracted from the NSW Registry of Births, Deaths, and Marriages registration file (RBDM) and the Cause of Death Unit Record File (COD).

Data linkage was performed by the NSW Ministry of Health Centre for Health Record Linkage and the Australian Institute of Health and Welfare (AIHW) Data Integration Services Centre. This study was granted ethical approval by the University of New South Wales, NSW Population and Health Services Research (HREC/18/CIPHS/56), Aboriginal Health and Medical Research Council of NSW (1503/19), and Australian Institute of Health and Welfare (EO2018/2/431) research ethics committees. AIHW privacy regulations require cell sizes of five or less to be suppressed due to risk of re-identification of individuals in the study.

## 2.2 Study cohort

The cohort included patients with a hospital episode with a primary diagnosis of AF or atrial flutter and a catheter ablation procedure in the same episode between January 2009 and December 2018 (codes in Appendix 2). The first such episode was identified as the index ablation. For each patient, we used three years of medical history before the index ablation and a maximum follow-up period of three years. Exclusion criteria included patients under 18 years of age at the time of index ablation, patients with a diagnosis of valvular heart disease or mitral valve stenosis, or a replacement of mitral valve procedure before or during the index ablation episode. If a hospital stay was made up of various episodes of care, we aggregated all diagnosis and procedure codes from these episodes into a single hospital stay (see Appendix 1).

## 2.3 Outcomes

We built prognostic models to predict common adverse outcomes for AF, with the ICD-10 codes used for each outcome based on prior observational studies (Appendix 2). We used a composite outcome of death, heart failure, stroke, and cardiac arrest, due to the low number of adverse events in the cohort. Major bleeding events (including gastrointestinal bleeding, intracranial bleeding, and other bleeding) were modeled separately and not included in the composite outcome since they tend to be a result of treatment and have different causes. Outcomes were obtained from the primary diagnosis from inpatient (APDC) and emergency records (EDDC), the death registry, and the cause of death records. The codes for each outcome are provided in Appendix 2.

## 2.4 Prognostic Machine Learning Algorithms

Survival models were built for the composite outcome and for major bleeding events. The composite outcome was modeled using single risk models, with censoring occurring at the earlier of three years of follow-up or December 2018. Major bleeding events were modeled using competing risks, with the composite outcome as the competing risk. The following survival algorithms were selected based on commonly used algorithms for prognosis and more recent deep survival models (further details in Appendix 3): Cox proportional hazards model with elastic net penalty (Cox); random survival forest (RSF); gradient boosted survival (GBT); DeepSurv, a Cox proportional hazards neural network (22); and deep survival machines (DSM), a neural network that estimates the survival function as a mixture of individual parametric survival distributions (Weibull or Lognormal) (23). For major bleeding events, DSM supports competing risks, but cause-specific modeling was used for all other algorithms. A cause-specific DSM was included for performance comparison. For the major bleeding cause-specific models, censoring was the earlier of experiencing the composite outcome, three years of follow-up, or December 2018. The $CHA_2DS_2$-VASc and HAS-BLED scores were included as baselines for the composite outcome and major bleeding events, with the risk estimates obtained from a cohort study of 182,678 AF patients (24).

## 2.5 Feature Extraction

All available diagnosis codes from inpatient episodes and all pharmaceutical data during the three-year lookback period and during the index ablation episode were used as features for the ML models. The features were coded in binary, with 1 indicating having experienced an event. Rare ICD-10-AM codes and medications associated with less than ten people were excluded. Diagnosis codes from emergency department visits were not used as features because the diagnosis coding in emergency administrative data contained multiple coding schemes. Sex and age at the time of the index ablation episode were included as features.

## 2.6 Evaluation

The prognostic models were evaluated using single-time-point metrics at the event horizon quantiles of 25%, 50%, and 75% for risk discrimination (using the time-dependent concordance index) and calibration (using the expected $\ell_1$ calibration error [ECE]). ECE measures the absolute difference between the observed and expected event rates, conditional on the estimated risk scores (25), with the expected event rates estimated using a Kaplan-Meier (KM) curve. These single-time-point metrics were adjusted with an inverse propensity of censoring estimate. Calibration over the entire distribution of the prognostic models was assessed with distributional calibration (D-calibration), with the D-calibration test indicating that the survival curve generated by the model for patients is not calibrated if P-value < 0.05 (26). KM curves for the composite outcome and major bleeding events were generated to estimate the cohort average cumulative risk of experiencing these events.

For each model, we performed 10-fold cross-validation, reporting the mean and standard deviation across the folds. The best model for each fold was selected by performing hyperparameter tuning on each fold (see Appendix 3). Post-hoc explanations of the best performing models were obtained with SHapley Additive exPlanations (SHAP), using the Kernel SHAP method for the deep survival models and the Tree SHAP method for the survival tree ensemble models (27,28), and visualized with summary plots. In this paper, SHAP explains the risk prediction of a patient by calculating the contribution of each feature to the risk prediction. SHAP summary plots show the magnitude and direction of feature attributions to the risk prediction, with the SHAP values from each validation fold aggregated to generate the summary plots.

## 3 Results

### 3.1 Patient characteristics

A total of 3285 patients had a hospitalization with AF as the primary diagnosis and an ablation procedure performed during that episode of care. The median age was 63 years (interquartile range, 56 – 70 years) and 1110 (33.8%) were female (Table 1). All patients were followed for a median of three years, patients who experienced the composite outcome—heart failure, stroke, cardiac arrest, and death—had a median follow-up of 10 months, and patients who experienced major bleeding events had a median follow-up of 12 months. The incidence of adverse events was low for all outcomes: composite (177, 5.4%), major bleeding (167, 5.1%), heart failure (103, 3.1%), stroke (18, 0.5%), cardiac arrest (0.2%), and death (75, 2.2%). The highest rate of events for both the composite outcome and major bleeding events occurred within the first five months of follow-up (KM plots in Appendix 4). At baseline, patients who experienced the composite outcome had higher $CHA_2DS_2$-VASc scores and higher rates of medical history of heart failure, hypertension, diabetes, and vascular diseases (Table 1). Patients who experienced major bleeding events had higher $CHA_2DS_2$-VASc scores and higher rates of medical history of hypertension (Table 1). Amongst those who died during follow-up, half had a history of hypertension.

### 3.2 Prognostic Models for Composite Outcome and Major Bleeding Events

Table 2 shows the prognostic model performance for predicting the composite outcome. GBT achieved the highest concordance index on the 25% and 75% quantile of event time horizons. The Cox and DeepSurv models were competitive at the 50% quantile of event time horizons. DeepSurv had the lowest expected calibration error on all three quantiles of event time horizons. For the composite outcome, the concordance index of the Cox model and the survival ensembles (GBT and RF) was similar, with GBT having slightly better performance on the 50% and 75% quantiles of event times. The calibration scores suggest that the estimated risks of the adverse outcomes are consistent with the cumulative probabilities of the KM curve. All models except for DSM were D-calibrated. The low number of events made visual assessment of calibration unreliable, but calibration plots of GBT at the event quantiles show that the model increasingly underestimates the risk of experiencing the event at the longer event horizons (Appendix 5). The $CHA_2DS_2$-VASc score had better risk discrimination performance at the 25% quantile of event times, but the ML survival models provided better discrimination performance on longer horizons (the 50% and 75% quantiles of event times). The ML models also had better calibration performance across all three quantiles of event times.

Table 3 shows the prognostic model performance for predicting major bleeding events. The risk discrimination performance of all models was low, with only DSM achieving a concordance index greater than 0.60 on all three quantiles of event times. DSM was the only model that was not D-calibrated. The cause-specific DSM model performance was on par with the other cause-specific models, which was lower than the competing risk DSM. The HAS-BLED also had poor discrimination (concordance index < 0.60) and calibration performance.

### 3.3 Explainability

For the composite outcome (Figure 1), older age, and a medical history of heart failure (congestive heart failure and left ventricular failure), fluid overload, disorders of magnesium metabolism, atherosclerotic heart disease, pneumonia, long term (current) use of anticoagulants, primary hypertension, presence of a cardiac device, and chronic kidney disease contributed to the model predicting higher risk. A medication history of furosemide (high-ceiling diuretic), spironolactone (aldosterone antagonist), cefalexin (antibacterial), amiodarone (antiarrhythmic), metformin (blood glucose lowering), warfarin (antithrombotic), bisoprolol (beta blocking agent), and allopurinol (antigout preparation) contributed to the model predicting higher risk. A medication history of flecainide (antiarrhythmic) contributed to lower risk.

Figure 2 shows the post-hoc explanation of the DSM model for predicting major bleeding events. Age had the greatest average impact on the model predictions, with older age contributing to a higher risk prediction. A medical history of tobacco use (past and present) and primary hypertension contributed to higher risk. A diagnosis code for injury, poisoning or other adverse effect with place of occurrence recorded as a health service area also contributed to the model predicting higher risk. A medication history of selected antibacterials and antibiotics (amoxicillin and beta-lactamase inhibitor, cefalexin, chloramphenicol, amoxicillin), warfarin (antithrombotic), atorvastatin (lipid modifying agent), pantoprazole (proton pump inhibitor), allopurinol (antigout preparation), amiodarone (antiarrhythmic), and prednisone (corticosteroid) contributed to higher risk. A medical history of paroxysmal AF, chronic hypertension, unspecified AF and atrial flutter, and a medication history of flecainide (antiarrhythmic) and class III antiarrhythmics contributed to predicting lower risk.

# 4 Discussion

This study evaluated survival prognostic ML models for predicting adverse outcomes in patients following catheter ablation treatment. The main findings of this study are: (1) patients in our cohort who underwent catheter ablation experienced adverse outcomes at low rates (5.7% for composite—death, heart failure, stroke, and cardiac arrest—and 5.3% for major bleeding events); (2) we succeeded in predicting experiencing the composite outcome with high precision, but prediction of experiencing major bleeding events was poor; (3) the most important features for predicting higher risk of experiencing the composite and the major bleeding outcomes were indicative of sicker and older patients.

Our study is the first to use ML-based survival models to predict adverse outcomes on patients following catheter ablation. Prior studies have focused on using risk scores that use a small number of variables (six or less) for predicting AF recurrence in patients following catheter ablation (19). Other studies have explored ML for predicting adverse events after catheter ablation, including AF recurrence (12–14) and 30-day and 90-day readmission (16,17). Our study improves on prior literature by predicting the risk of major cardiovascular events post catheter ablation, which are the primary endpoints that have been assessed when determining the effectiveness of catheter ablation over pharmacotherapy (8,9,29).

## 4.1 Meaning of study

The strong performance results for predicting the composite outcome highlight its potential clinical utility. The competitive performance in complex models (gradient boosted survival) and interpretable models (elastic net Cox) demonstrates that various deployment options are available to clinical teams, depending on the prioritization of transparency, maximal performance gains, or calibration. The poor performance of models predicting major bleeding events suggests: (1) that the composite outcome is easier to model than major bleeding events; (2) that diagnosis history and prescription claims prior the catheter ablation treatment do not contain sufficient information to predict future major bleeding events; (3) future research should explore additional clinical and biological variables for predicting major bleeding events.

Older age was the factor that had the highest impact on predicting risk of experiencing the composite outcome and major bleeding events. A medication history of flecainide contributed to predicting a lower risk of experiencing the composite outcome and major bleeding events, which may be due to flecainide being selectively used in lower risk patients (as recommended in clinical guidelines for the management of AF (5)). Known cardiovascular risk factors and markers of underlying disease contributed to predicting higher risk, including hypertension, heart failure, heart disease, presence of a cardiac device, and tobacco use. This is consistent with factors identified in a prior study that used deep learning the risk of experiencing CVD events (30). Comorbidities indicative of poor health (pneumonia, chronic kidney disease), therapies used in sicker patients to treat heart failure and AF (antiarrhythmics, antithrombotics, diuretics, and beta blocking agents) also contributed to higher risk predictions of adverse outcomes. A health service area coded as the place of occurrence for an external cause code is indicative of healthcare-related adverse events and complications, including medication errors, which may have occurred prior to or during a hospital stay. The best model for the composite outcome had more comorbidities in the top 20 features, whereas the best model for major bleeding had more antibacterials and antibiotics, suggesting complications and sicker patients. Paroxysmal AF is associated with better prognosis than permanent AF, reflected by a diagnosis of paroxysmal AF contributing to lower risk prediction.

The models developed in this study provide an example of prognostic modeling that could be incorporated in clinical practice following the administration of catheter ablation. Identifying patients at high risk may lead to proactive management by clinicians.

### 4.2 Machine Learning Implications

For the composite outcome, the deep survival models did not outperform the traditional survival algorithms. This may be due to the censoring rate (> 94%), the low number of events, and the tabular dataset (where deep learning is not guaranteed to outperform other models). For major bleeding events, the difference in performance between the cause-specific DSM and the competing risk DSM suggests that the representation learning layer for the competing risks captures additional information that leads to significant performance gains (23).

Caution must be taken with the interpretation of the summary plots and the impact of the features on risk predictions. The SHAP explanations capture correlations between the input features and the risk prediction, but they are not causal. While explanation of the models focused on the top 20 features, other features not listed also impacted the risk predictions. Explanations with SHAP are particularly useful when paired with models that use non-linearities and capture complex interactions in the input features, such as the survival ensembles and deep survival models.

### 4.3 Limitations

The high degree of censoring and low number of events in our cohort may limit how well our results generalize. This study did not include data regarding family history, outpatient medical history, social and lifestyle factors (except in cases where tobacco use was recorded in a hospital admission). Our analysis relied on coded hospital diagnoses, recorded only for conditions that significantly affect patient management during an episode of care. Potential sources of bias in our analysis include using diagnoses as a proxy for incident or prevalent disease and the quality of hospital diagnosis coding (accuracy 51-98% across 32 studies) (31). Given the high rate of history of heart failure amongst patients who experienced the composite outcome (34.5%) and history of heart failure being an important feature for predicting the composite outcome, a separate study on a heart failure cohort would be desirable, but it is left for future work due to the small size of our current dataset.

### 4.4 Conclusion

ML survival models using diagnostic and medication history predicted the risk of experiencing a composite adverse outcome—death, heart failure, stroke, and cardiac arrest— with high precision following catheter ablation for AF. Using the same data and ML survival models could not predict major bleeding events. The models presented may be useful in proactively managing high-risk patients following catheter ablation. Future research is needed to validate the usefulness of these models in clinical practice.

## 5 Conflict of interest

The authors declare that the research was conducted in the absence of any commercial or financial relationships that could be construed as a potential conflict of interest.

## 6 Author contributions

All authors contributed to the conception and design of the study. JCQ performed the data analysis and wrote the first draft. All authors contributed to critical revisions of the manuscript. All authors approved the final draft.

## 7 Funding

This work was supported by National Health and Medical Research Council (NHMRC), project grant No. APP1184304. Creation of the linked dataset was funded by a NHMRC Project Grant No. 1147430.This work was supported by National Health and Medical Research Council (NHMRC), project grant No. APP1184304. Creation of the linked dataset was funded by a NHMRC Project Grant No. 1147430.

## 8 Acknowledgements

**Data Availability Statement**

The datasets for this study were obtained from the New South Wales (NSW) Ministry of Health Centre for Health Record Linkage and the Australian Institute of Health and Welfare (AIHW) Data Integration Services Centre. The datasets cannot be publicly shared due to ethical, governance, and confidentiality agreements.

1   Table 1 - Summary of baseline characteristics of patients at time of index catheter ablation for atrial fibrillation (AF) admission to a public or
2   private hospital in New South Wales, Australia.

| | All | Composite * | Major Bleeding | HF | Stroke | Cardiac Arrest | Death |
|---|---|---|---|---|---|---|---|
| N | 3285 | 177 | 167 | 103 | 18 | 7 | 75 |
| **Gender** | | | | | | | |
| Male | 2175 (66.2) | 108 (61.0) | 115 (68.9) | 62 (60.2) | 13 (72.2) | < 5 | 43 (57.3) |
| Female | 1110 (33.8) | 69 (39.0) | 52 (31.1) | 41 (39.8) | 5 (27.8) | < 5 | 32 (42.7) |
| **Age, median (Q1, Q3)** | 63, (56, 70) | 71 (64, 76)† | 66 (58, 71)† | 70 (64, 75.5)† | 70 (62, 76)† | 64 (52, 74) | 72 (63.5, 77)† |
| <65 | 1767 (53.8) | 50 (28.2) | 76 (45.5) | 27 (26.2) | 6 (33.3) | < 5 | 21 (28.0) |
| 65-<75 | 1171 (35.6) | 74 (41.8) | 65 (38.9) | 47 (45.6) | 6 (33.3) | < 5 | 29 (38.7) |
| >=75 | 347 (10.6) | 53 (29.9) | 26 (15.6) | 29 (28.2) | 6 (33.3) | < 5 | 25 (33.3) |
| **Follow-up months, median (Q1, Q3)** | 36 (20, 36) | 9 (3, 22) † | 12 (5, 23)† | 8 (2, 20) † | 14 (7, 24)† | 7 (3, 19)† | 10 (3, 22) † |
| **CHA$_2$DS$_2$-VASc, median (Q1, Q3)** | 1 (0, 2) | 3 (2, 3) † | 2 (1, 2.5)† | 3 (2, 3.5)† | 2 (1, 3)† | 2 (1, 3.5) | 3 (2, 4) † |
| Congestive heart failure | 277 (8.4) | 61 (34.5) † | 21 (12.6) | 37 (35.9) † | < 5 | < 5 | 29 (38.7) † |
| Hypertension | 816 (24.8) | 77 (43.5) † | 60 (35.9) † | 43 (41.7) † | 7 (38.9) | < 5 | 37 (49.3) † |
| Diabetes | 388 (11.8) | 42 (23.7) † | 19 (11.4) | 26 (25.2) † | < 5 | < 5 | 17 (22.7) † |
| Stroke or TIA | 113 (3.4) | 6 (3.4) | 5 (3.0) | < 5 | < 5 | 0 | < 5 |
| **Vascular diseases** | 141 (4.3) | 18 (10.2) † | 10 (6.0) | 14 (13.6) † | < 5 | < 5 | 7 (9.3) † |

* Composite: Death, heart failure, cardiac arrest, and stroke. **Vascular diseases: coronary artery disease, peripheral artery disease, atherosclerosis, and myocardial infarction. Values are given as median and IQR, or total number (n) and %. †P-value < 0.05. P-values for continuous variables are the results of the Mann-Whitney U test on patients who experienced the event and those who did not experience the event. For categorical variables, Fisher's Exact test was used to compare frequencies.

Table 2 – Concordance index and expected calibration error results for the machine learning survival models that predict the composite outcome of death, heart failure, cardiac arrest, and stroke. Standard deviation from the 10-fold cross-validation shown in parenthesis.

| Models | Concordance Index | | | Expected Calibration Error | | | D-cal |
|---|---|---|---|---|---|---|---|
| | Quantiles of Event Times | | | Quantiles of Event Times | | | |
| | 25% | 50% | 75% | 25% | 50% | 75% | |
| Cox | 0.80 (0.08) | 0.81 (0.08) | 0.77 (0.08) | 0.016 (0.002) | 0.027 (0.003) | 0.037 (0.006) | 0.999 |
| RSF | 0.77 (0.13) | 0.79 (0.09) | 0.78 (0.07) | 0.017 (0.003) | 0.026 (0.006) | 0.037 (0.009) | 0.999 |
| GBT | 0.82 (0.10) | 0.80 (0.08) | 0.79 (0.06) | 0.021 (0.003) | 0.034 (0.008) | 0.048 (0.012) | 0.999 |
| DeepSurv | 0.80 (0.13) | 0.80 (0.11) | 0.78 (0.10) | 0.014 (0.003) | 0.020 (0.006) | 0.028 (0.007) | 0.999 |
| DSM | 0.77 (0.16) | 0.76 (0.12) | 0.76 (0.10) | 0.021 (0.006) | 0.033 (0.010) | 0.046 (0.016) | 0.004 |
| $CHA_2DS_2$-VASc | 0.83 (0.09) | 0.78 (0.07) | 0.74 (0.09) | 0.022 (0.007) | 0.043 (0.009) | 0.096 (0.021) | 0.004 |

Table 3 – Concordance index and expected calibration error results for the machine learning survival models that predict major bleeding events. Standard deviation from the 10-fold cross-validation shown in parenthesis.

| Models | Concordance Index | | | Expected Calibration Error | | | D-cal |
|---|---|---|---|---|---|---|---|
| | Quantiles of Event Times | | | Quantiles of Event Times | | | |
| | 25% | 50% | 75% | 25% | 50% | 75% | |
| Cox | 0.53 (0.16) | 0.59 (0.07) | 0.57 (0.05) | 0.029 (0.011) | 0.040 (0.026) | 0.051 (0.024) | 0.999 |
| RSF | 0.54 (0.11) | 0.59 (0.09) | 0.59 (0.08) | 0.018 (0.002) | 0.025 (0.003) | 0.034 (0.008) | 0.999 |
| GBT | 0.57 (0.12) | 0.58 (0.10) | 0.58 (0.06) | 0.018 (0.003) | 0.025 (0.004) | 0.029 (0.005) | 0.999 |
| DeepSurv | 0.57 (0.08) | 0.57 (0.10) | 0.57 (0.09) | 0.018 (0.002) | 0.026 (0.003) | 0.032 (0.006) | 0.999 |
| DSM – cause specific | 0.54 (0.10) | 0.55 (0.10) | 0.55 (0.09) | 0.022 (0.005) | 0.033 (0.008) | 0.041 (0.014) | 0.225 |
| **DSM – competing risks** | **0.66 (0.09)** | **0.64 (0.08)** | **0.61 (0.08)** | **0.023 (0.007)** | **0.033 (0.008)** | **0.046 (0.012)** | **< 0.001** |
| HAS-BLED | 0.54 (0.06) | 0.56 (0.09) | 0.58 (0.07) | 0.068 (0.009) | 0.186 (0.017) | 0.359 (0.033) | < 0.001 |

**FIGURE LEGENDS**

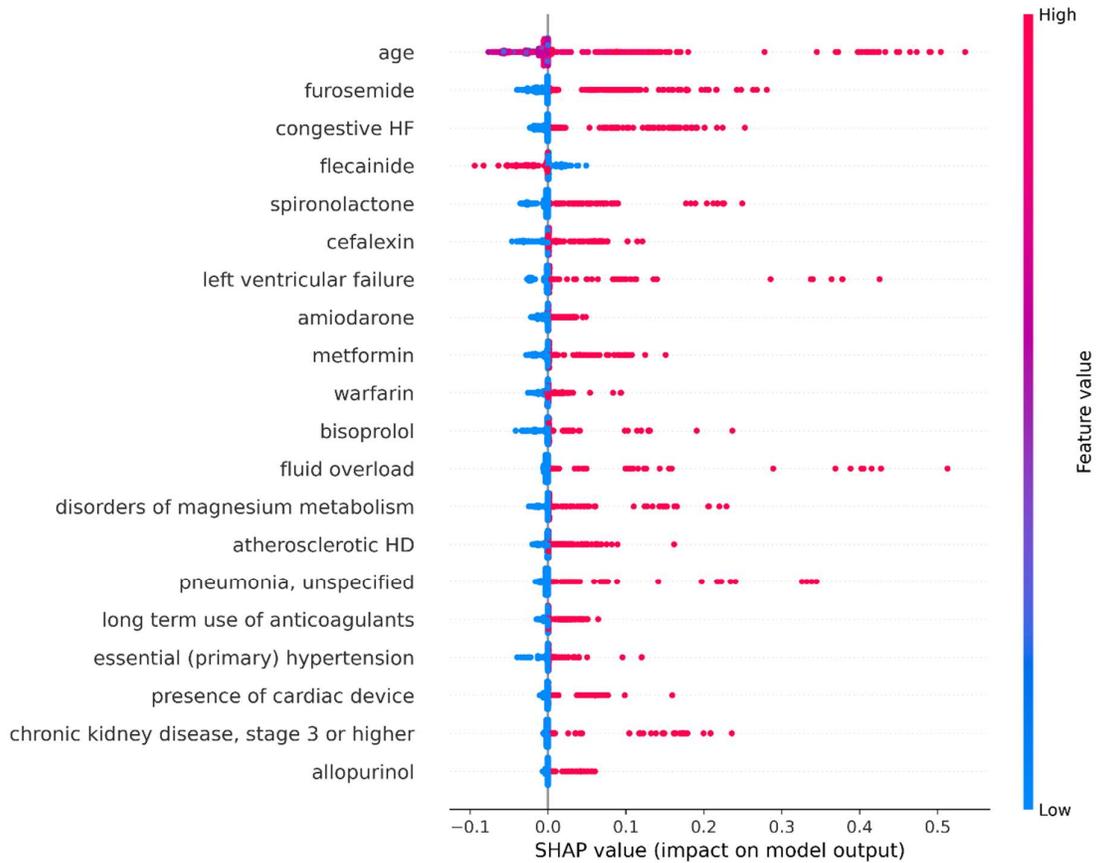

Figure 1. SHAP summary plot for the best performing model (gradient boosting survival) for predicting risk of experiencing the composite outcome of death, heart failure, cardiac arrest, and stroke, calculated on predictions made on the 25% quantile of time events. The set of beeswarm plots indicate the impact of the top 20 features on the model output. Each dot represents an individual in the dataset. The position of the dot on the x axis represents the impact that feature has on the model's prediction for that individual. The density represents multiple dots being at the same position. Binary features are represented by two colours: 1 – high/red and 0 – low/blue).

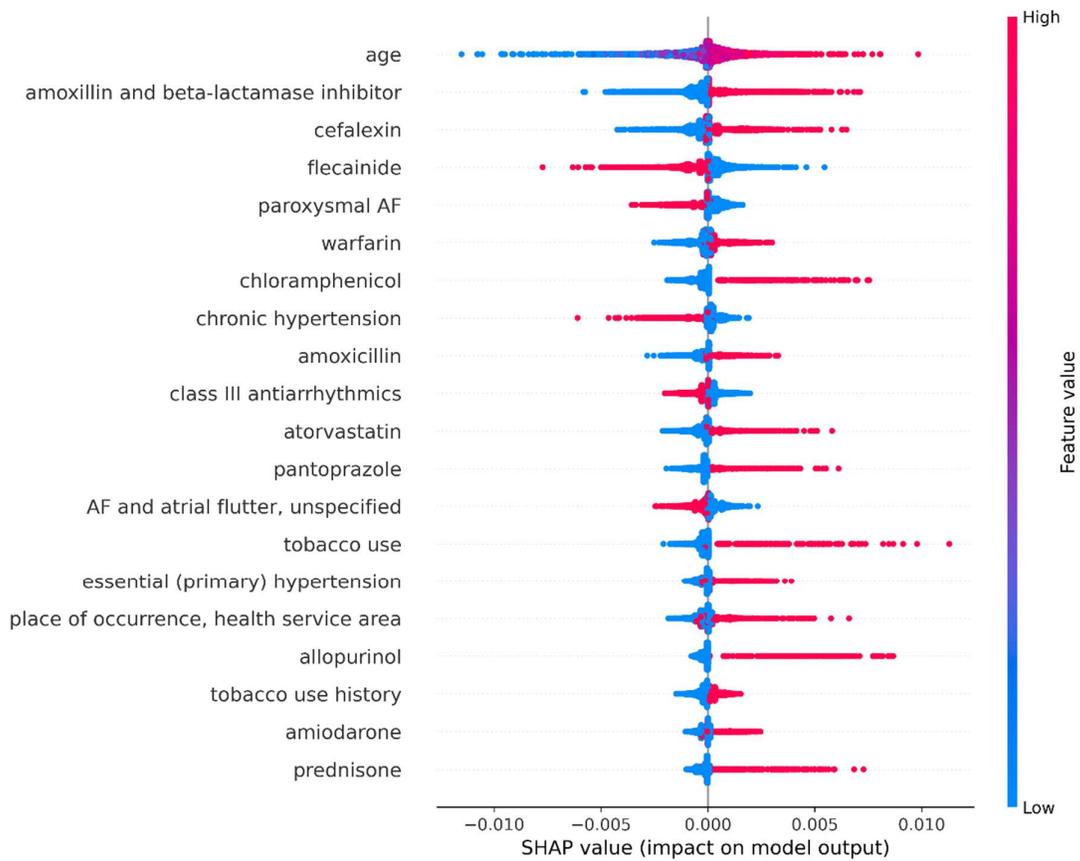

Figure 2. SHAP summary plot for the best performing model (deep survival machine) for predicting risk of experiencing major bleeding events, calculated on predictions made on the 25% quantile of time events. The set of beeswarm plots indicate the impact of the top 20 features on the model output. Each dot represents an individual in the dataset. The position of the dot on the x axis represents the impact that feature has on the model's prediction for that individual. The density represents multiple dots being at the same position. Binary features are represented by two colours: 1 – high/red and 0 – low/blue).

## 2.1 Appendix 1

The New South Wales (NSW) Admitted Patient Data Collection (APDC) includes records of all inpatient separations (discharges, transfers, and deaths) from public and private hospitals in NSW, Australia. The APDC records up to 50 procedures, coded using the Australian Classification of Health Interventions (ACHI) and up to 51 diagnoses, coded using the International Classification of Diseases, 11th Revision, Australian Modification (ICD-10-AM).

Admitted patient records relate to episodes of care, defined as the period of admitted patient care between a formal or statistical admission and a formal or statistical separation, characterised by only one care type. A "statistical" admission or separation records the commencement or cessation of an episode of care, which may occur when there is a change in the type of care provided to a patient (e.g. from acute to subacute care). A "formal" admission or separation records the commencement or cessation of a patient's treatment and/or care and/or accommodation.

A hospital stay, defined as the period of admitted patient care between a formal admission and a formal discharge, may comprise one or more episodes of care. For episodes of care ending in type change (e.g. from acute to sub-acute care) or transfer, we constructed contiguous periods of stay using admission dates and admission status from the first episode and separation dates and separation type from the last episode of care. Therefore, hospital stays as defined in this report include continuous periods of inpatient care that were provided by one or more hospitals.

Diagnoses in the NSW Emergency Department Data Collection (EDDC), are coded using SNOMED CT, ICD9CM, ICD10V8, and ICD-10-AM.

## 2.2 Appendix 2

**Outcome Codes**

The codes used for each outcome are based on prior observational studies (1,2). Diagnosis were coded using the International Classification of Diseases, 11th Revision, Australian Modification (ICD-10-AM). Procedures were coded using the Australian Classification of Health Interventions (ACHI).

| Outcome | ICD-10-AM Codes | | ICD 9 Codes |
|---|---|---|---|
| Heart failure hospitalization | I50.x, I11.0, I13.0, I13.2 | | 428 |
| Ischemic stroke | I63.X | | 433.x1, 434.x1, 436 |
| Major bleeding | | | 456.0, 456.20, 530.21, 530.7, 530.82, 531.[0246], 532.[246], 533.[0246], 534.[246], 534.[246], 535.[0-7]1, 537.8[34], 562.0[23], 562.1[23], 569.3, 569.85, 578, 430, 431, 432, 85[23], 800.[2378], 801.[2378], 803.[2378], 804.[2378], 423.0, 459.0, 568.81, 569.7, 599.71, 719.1, 784.8, 786.3 |
| | Gastrointestinal bleeding | I85.0, K22.1, K22.6, K25.0, K25.2, K25.4, K25.6, K26.0, K26.2, K26.4, K26.6, K27.0, K27.2, K27.4, K27.6, K28.0, K28.4, K28.6, K29.x1, K31.82, K55.22, K57.x1, K57.x3, K62.5, K92.0, K92.1, K92.2, | |
| | Intracranial bleeding | I60.x, I61.x, S06.34, S06.35x, S06.36x, S06.37x, S06.38x, S06.4, S06.5, S06.6 | |
| | Other bleeding | I31.2, K66.1, M25.0, R04.1, R04.2, R31, R58 | |
| Cardiac arrest | I46 | | 427.5 |

**Inclusion and Exclusion Criteria Codes**

| Criteria | Codes | | Coding Scheme |
|---|---|---|---|
| Atrial fibrillation or atrial flutter | I48.x | | ICD-10-AM |
| Catheter ablation | 38290-01, 38287-02 | | ACHI |
| Valvular heart disease | | | ICD-10-AM |
| | Rheumatic mitral valve diseases | I05.x | |
| | Congenital malformations of aortic and mitral valves | Q23.x | |
| Mitral valve stenosis | Nonrheumatic mitral valve disorders | I34.x | ICD-10-AM |
| Replacement of mitral valve procedure | 38488-09, 38488-02, 38488-03, 38489-02 | | ACHI |

For HAS-BLED and $CHA_2DS_2$-VASc ICD-10 codes and ATC codes see (3).

## 2.3 Appendix 3

All models were implemented in Python. Hyperparameter optimization was implemented with Optuna with 20 trials using three-fold cross-validation on the training set. Cox proportional hazards model with elastic net penalty (1), random survival forest (3), and gradient boosted survival (4) used the implementation from the scikit-survival library (2). The neural network survival models DeepSurv (5) and deep survival machines (DSM) (7) used the implementation from the auton-survival library (6). The neural networks were trained for 10 epochs, with the number of epochs chosen by experimentation on a subset of the data.

**Table 4 – Hyperparameter values for Cox proportional hazards with elastic net penalty.**

| Parameter | Value ranges |
|---|---|
| Alpha | Uniform(0.0001, 0.05) |
| L1 ratio | {0.2, 0.3, 0.4, 0.5, 0.6, 0.7, 0.8, 0.9, 1} |

**Table 5 – Hyperparameter values for random survival forest.**

| Parameter | Value ranges |
|---|---|
| number of estimators | {10, 50, 100, 200} |
| max depth | {3, 5, 10, None} |
| max features | {'sqrt', 'log2', 100, 200} |
| min samples leaf | {3, 10, 150, 200} |

**Table 6 – Hyperparameter values for gradient boosted survival.**

| Parameter | Value ranges |
|---|---|
| number of estimators | {10, 50, 100, 200} |
| max depth | {3, 5, 10} |

| max features | {'sqrt', 'log2', 100, 200} |
| min samples leaf | {1, 3, 10, 150, 200} |
| learning rate | {0.001, 0.01, 0.1, 0.5} |

Table 7 – Hyperparameter values for DeepSurv.

| Parameter | Value ranges |
| --- | --- |
| layers | {1, 2, None} |
| nodes per layer | {10, 20, 50, 100, 200} |
| learning rate | {0.0001, 0.001, 0.01} |
| batch size | {64, 128} |

Table 8 – Hyperparameter values for deep survival machines.

| Parameter | Value ranges |
| --- | --- |
| distribution | {'Weibull', 'LogNormal'} |
| k – number of underlying parametric distributions | {2, 3, 4, 5, 6} |
| discount | {0.5, 0.75, 1} |
| layers | {1, 2, None} |
| nodes per layer | {10, 20, 50, 100, 200} |
| learning rate | {0.0001, 0.001, 0.01} |
| batch size | {64, 128} |

## 2.4 Appendix 4

**Composite outcome**

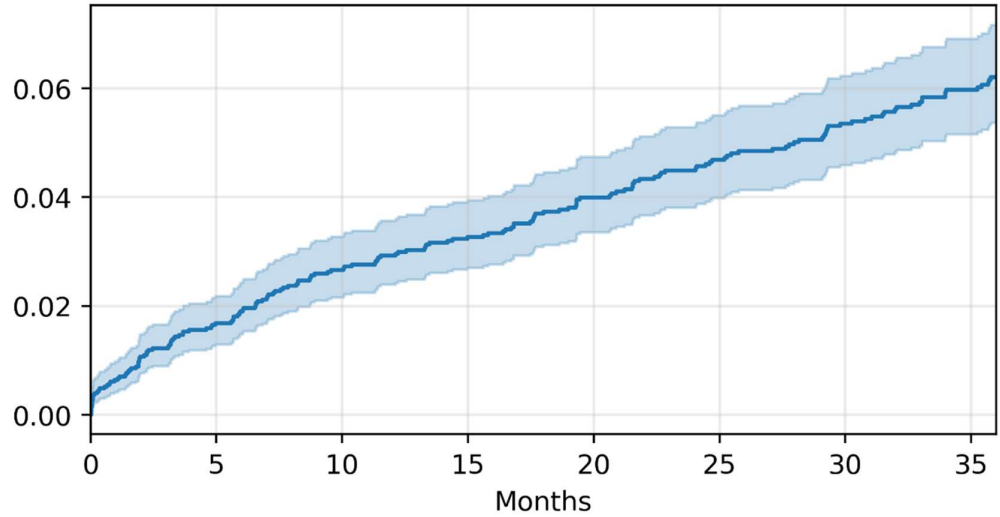

Figure 3 - Kaplan Meier survival curve showing the cumulative probability of experiencing the composite outcome—death, heart failure, cardiac arrest, stroke—after catheter ablation treatment for atrial fibrillation. The highest rate of events for the composite outcome occurred within the first five months of follow-up.

**Major bleeding**

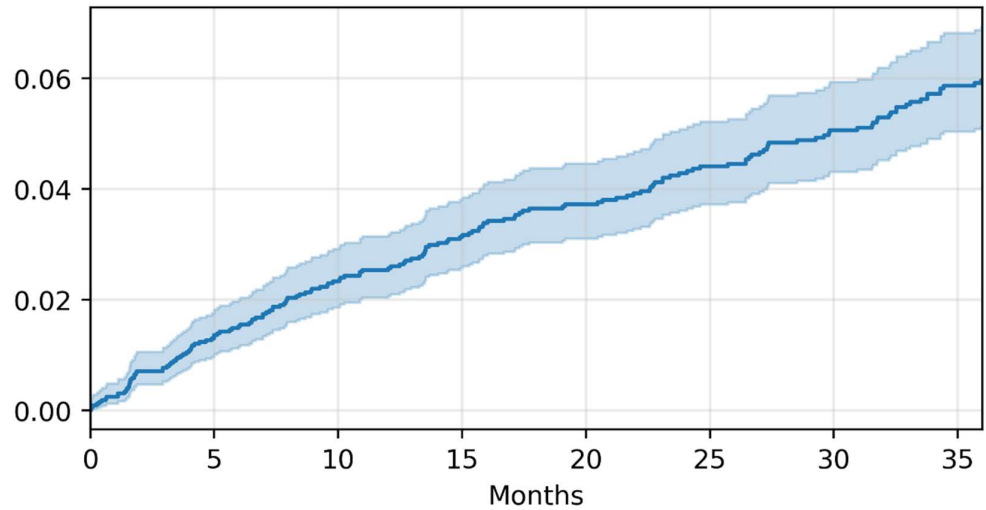

Figure 4 - Kaplan Meier survival curve showing the cumulative probability of experiencing major bleeding events after catheter ablation treatment for atrial fibrillation. The highest rate of events for the composite outcome occurred within the first five months of follow-up.

## 2.5 Appendix 5

Supplementary figures 3-5 show the calibration curves for gradient boosted survival for predicting the composite outcome of death, heart failure, cardiac arrest, and stroke. The figures illustrate the calibration curves at the three evaluated time horizons of 25%, 50%, and 75% of observed event times. The calibration curves were calculated as described in [2] using the implementation from the lifelines Python library [3], but restricting the range of predicted probabilities from the first to the 99th percentiles of risk in the population. Calibration curves are not included for deep survival machines for predicting major bleeding events because of the poor risk discrimination performance.

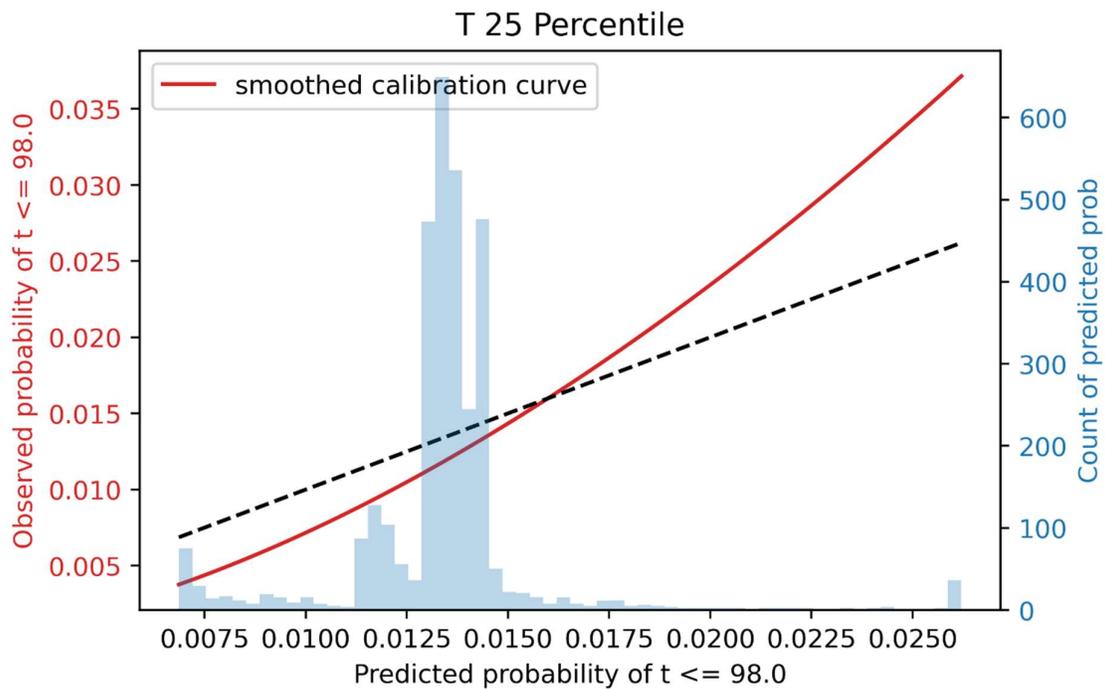

Figure 5 – Survival calibration curve at the first quantile (25%) of the event times for the gradient boosted survival model predicting the composite outcome of death, heart failure, cardiac arrest, and stroke.

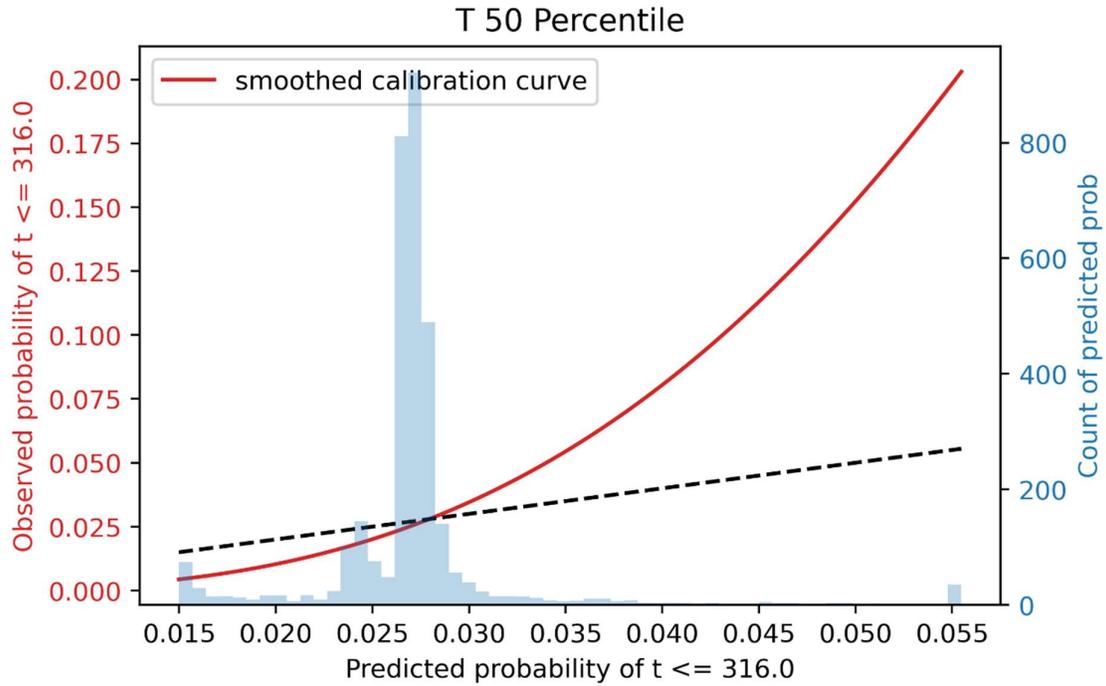

Figure 6 – Survival calibration curve at the 50% quantile of the event times for the gradient boosted survival model predicting the composite outcome of death, heart failure, cardiac arrest, and stroke.

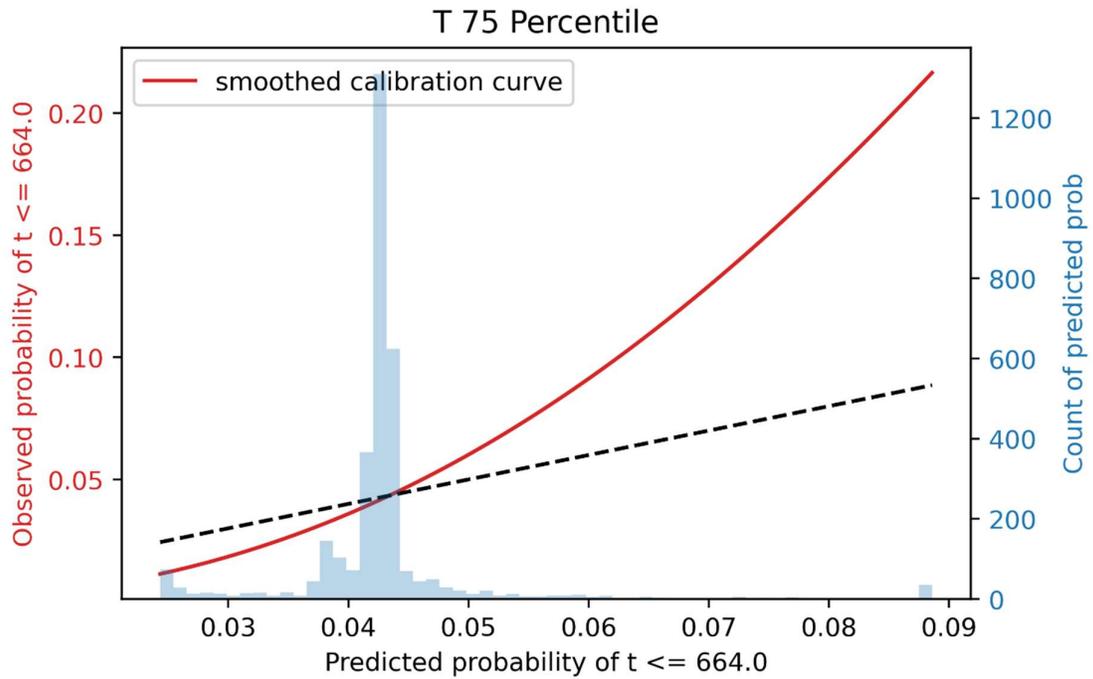

Figure 7 – Survival calibration curve at the 75% quantile of the event times for the gradient boosted survival model predicting the composite outcome of death, heart failure, cardiac arrest, and stroke.